\documentclass{article}

     \usepackage[preprint]{neurips_2020_tda_}

\usepackage{natbib}

\usepackage{subcaption}
\usepackage[utf8]{inputenc} %
\usepackage[T1]{fontenc}    %
\usepackage{amsfonts}       %
\usepackage{booktabs}       %
\usepackage{hyperref}       %
\usepackage{url}            %
\usepackage{microtype}      %
\usepackage{nicefrac}       %
\usepackage{paralist}       %
\usepackage{siunitx}        %
\usepackage{cmap}
\usepackage[linewidth=1pt]{mdframed}
\usepackage[all]{xy}
\usepackage{graphicx}
\usepackage{float}
\usepackage{amsmath,amssymb,amscd,mathrsfs}

\usepackage{algpseudocode}
\usepackage{algorithm} 
\usepackage{authblk}

\newcommand{\rk}{\ensuremath{\text {rank\,}}}

\title{
  Betti numbers of attention graphs is all you really need
}

\author[1]{\bf{Laida Kushnareva}} 
\author[2]{\bf{Dmitri Piontkovski}}
\author[1]{\bf{Irina Piontkovskaya}}

\affil[1]{Huawei Noah’s Ark lab, Moscow, Russia}
\affil[2]{HSE University, Moscow, Russia}
\begin{document}

\maketitle

\begin{abstract}
 We apply methods of topological analysis to the attention graphs, calculated on the attention heads of the BERT model~(\cite{devlin2019bert}). Our research shows that the classifier built upon basic persistent topological features (namely, Betti numbers) of the trained neural network can achieve classification results on par with the conventional classification method. We show the relevance %
 of such topological text representation on three text classification benchmarks. For the best of our knowledge, it is the first attempt to analyze the topology of an attention-based neural network, widely used for Natural Language Processing.
\end{abstract}

\section{Introduction}

Modern Neural Networks embed data into a high-dimensional space. Moreover, each layer and even a layer part  can be considered as separate embedding, where the information about  interconnections of these separate embeddings is encoded by some  weighted directed graph.  In particular, one can apply various methods to investigate such graphs for  attention  heads in multi-headed attention models, such as BERT.

Conventionally, the BERT model is used for sentence classification by adding a softmax-based classification layer upon the output embedding. Instead, we propose to use a linear classifier built solely upon the persistent topological features (namely, the first two Betti numbers) without using any information about the order of tokens or to which particular token each weight relates. We have found that it provides a classification quality on par with the conventional classification method in numerous tasks.
Moreover, on some tasks such as linguistic acceptability and spam detection, our topological classifiers outperform the usual BERT-based classification. We conclude that the topology of the attention graphs of the trained BERT model contains enough information for solving considered classification tasks. The second outcome is, that the proposed text representation, based only on the first two Betti numbers of the attention graph, can solve the task having lower %
dimensionality than BERT embedding.

The attention graphs are built as follows.
Each attention head in the Transformer architecture calculates weights of each token in the sentence with respect to every other token, and the next level representation is constructed using these weights. The attention graph for each head is a complete digraph (with loops which appear when the token "pays attention" to itself) whose vertices are the tokens and the attention weights are the weights of the edges. %

In the classifiers, we use the BERT-based classification model, which is initialized with pre-trained BERT weights and then is fine-tuned for a given two-class classification task. After fine-tuning, we extract the persistent features of each head of this model for each data sample and then train a logistic regression classifier upon these features. 

Note that our results also confirm that different attention heads contain different amounts of information. These results are well aligned to previous works on BERT  (\cite{NIPS2019_9551}, \cite{journals/corr/abs-1906-04341}).  %

\section{Related work}

There are several recent  insights obtained by topological analysis of the neural representations of realistic datasets. The results of \cite{naitzat2020topology}  demonstrate that a deep neural network with $ReLU$ activation function tends to simplify the topology of the data from layer to layer, with the smallest Betti numbers on the output representations. Topological features are also shown to be efficient for predicting the generalization ability of the network, its efficiency and stability to adversarial examples (\cite{corneanu2020computing}, \cite{corneanu2019what}, \cite{rieck2018neural}). An overview of persistent topology methods, both practical and theoretical, with the focus on Artificial Neural Networks analysis, can be found in \cite{otter2017roadmap}, \cite{chazal2017introduction}. At the same time, while many researchers are focused on applications of topology to modern AI algorithms, there are efforts in the mathematical society to further expand the set of applicable methods (\cite{bergomi2019towards}, \cite{chowdhury2019path}, \cite{manin2020homotopy}). 

\section{Background}
\subsection{Topological background}

In our approach, 
we use the following two numerical attributes of an arbitrary  
graph $G = (E,V)$: the number $\beta_0$ 
of connected components and the number $\beta_1$ of independent cycles of $G$.
If one considers the graph $G$  as a simplicial complex, these numbers are equal to its Betti numbers.  Note that the Betti numbers $\beta_0$ and $ \beta_1$
of a graph filtration keep the whole information about the persistent homology barcodes, see Appendix A %
for details.

\subsection{BERT model}

BERT (\cite{devlin2019bert}) is the pre-trained model, which achieves state of the art results for many NLP tasks. The model is based on Transformer architecture, introduced in \cite{vaswani2017attention}. The BERT model is pre-trained on the large amount of data with Masked Language Modelling and Next Sentence Prediction objectives. For downstream tasks the task-specific classifier is attached to the BERT output layer and the model is fine-tuned. In our experiments we use the uncased BERT-base %
version, which consists of 12 layers, with 12 attention heads in each. The input of 
each attention head is a matrix $X$ consisting of  the $d$-dimensional representations (row-wise) of $m$ 
tokens of the sentence, so that $X$ is of size $m \times d$. The output of the head is the updated matrix of the representations $X^\mathrm{out}$, that is,
\begin{equation}
\begin{split}
X^\mathrm{out} & = W^\mathrm{attn}(XW^{\mathrm V}) \\
 \mbox{ with } 
 W^\mathrm{attn} & = \mathrm{softmax}\left(\frac{(XW^{\mathrm Q})(XW^\mathrm{K})^\mathrm{T}}{\sqrt{d}}\right),
\label{eqn:attention}
\end{split}
\end{equation}
where %
$W^\mathrm{Q}$, $W^\mathrm{K}$, $W^\mathrm{V}$ are trained projection matrices of size $d \times d$ and $W^\mathrm{attn}$ is the $m \times m$ 
matrix of attention weights (cf. \cite[Sec.~3.2]{vaswani2017attention}). 
One can interpret each element 
$w^\mathrm{attn}_{ij}$ 
as a weight of  $j$-th input's influence on   $i$-th output; larger weights mean stronger connection between corresponding tokens. 

\section{Our method} 
                  
Let us be given some dataset $S = \{s_i\}_{i=1}^N$ of $N$ natural language texts
encoded with $m$ tokens each and pre-trained attention-based model $M$. First of all, we fix some set of thresholds $T = \{t_i\}_{i=1}^{k}, 0 < t_1 < t_2 < ... < t_k < 1$ and chose a subset of heads of the model $H_M$, on which we will perform calculations.
       
Then we feed each text sample $s = s_i$ to the input of the model $M$ and obtain the matrix $W^\mathrm{attn} = (w^{attn}_{i, j})$ on each head $h \in H_M$. This matrix defines a weighted complete digraph with loops $\mathsf{\Gamma^{h}_s}$ with $m$ vertices, where $w^\mathrm{attn}_{ij}$ 
is the weight of the edge $j \to i$. 

After it, for each graph $\mathsf{\Gamma^{h}_s}$ and for each threshold level $t_i \in T$ we build an unweighted directed graph $\Gamma^{h}_s(t_i)$ as follows. The set of vertices of $\Gamma^{h}_s(t_i)$ is the same as the one for the graph $\mathsf{\Gamma^{h}_s}$, moreover, an edge of 
 $\mathsf{\Gamma^{h}_s}$ belongs to the new graph $\Gamma^{h}_s(t_i)$  if and only of its weight in $\mathsf{\Gamma^{h}_s}$  is at least  $t_i$. 
This way we assign a sequence of graphs $\Gamma^{h}_s(t_i), t_i \in T$ to each text sample for each head of the model.

For each unweighted directed graph $\Gamma^{h}_s(t_i)$ we also consider the corresponding 
undirected graph $\overline{\Gamma^{h}_s(t_i)}$ by setting an undirected edge $v_1v_2$ for each pair of vertices $v_1$ and $v_2$ 
which are connected by an edge in at least one direction in the graph $\Gamma^{h}_s(t_i)$. Then we count $\beta_0, \beta_1$ of undirected graph $\overline{\Gamma^{h}_s(t_i)}$. More precisely the process of features calculation for each data sample is described in Algorithm \ref{our_alg}.

\begin{algorithm}
	\caption{Topological features calculation} 
    \label{our_alg}
	\begin{algorithmic}[2]
    	\smallskip
	    \Require {Text sample $s$}
	    \Require {Set of chosen attention heads $H_M$ of attention-based model $M$}
	    \Require {Thresholds array $T$}
	    \smallskip
	    \Ensure {Features array $Features$}
		\medskip
		\Procedure{features\_calculation}{$s, H_M, T$}
	        \ForAll {$h \in H_M$}
		        \State Calculate attention graph $\mathsf{\Gamma^{h}_s} = (V, E, W^{attn}_{h, s})$ on sample $s$ on head $h$
		        \ForAll {$t \in T$} \Comment{Filtration:}
		            \State $E^h_s(t) \leftarrow \{e \in E(\mathsf{\Gamma^{h}_s}): W^{attn}_{h, s}(e) \geq t\}$  \Comment{Removing edges of weight less than $t$}
    				\State $\Gamma^{h}_s(t) \leftarrow (V, E^h_s(t))$ \Comment{Ignoring weights of remaining edges}
    				\State $\overline{E^h_s(t)} \leftarrow \left\{ \{i, j\} : (i, j) \in E^h_s(t) \right\} $ \Comment{Ignoring edges directions}
    				\State $\overline{\Gamma^h_s(t)} \leftarrow (V, \overline{E^h_s(t))}$ 
    				\State Calculate 
    				$\beta_0(\overline{\Gamma^h_s(t)}), \beta_1(\overline{\Gamma^h_s(t)})$ \Comment{Calculating Betti numbers of undirected graph}
    				
    			\EndFor
			\EndFor
			\State $Features \leftarrow \left[ \beta_0(\overline{\Gamma^h_s(t)}), \beta_1(\overline{\Gamma^h_s(t)})\right]_{t \in T}^{h \in H_M}$ %
    		\State \textbf{return} $Features$ 
    	\EndProcedure
	\end{algorithmic} 
\end{algorithm}

After this features calculation, we train the logistic regression on features, obtained for each sample of the train subset of the dataset and then make predictions on features of samples from the test subset.

\section{Experiments}

\subsection{Datasets}

We performed our experiments on the following datasets, labeled for different classification tasks.

\textbf{The Corpus of Linguistic Acceptability} ("CoLA") dataset (\cite{warstadt2018neural}) contains 10,657 sentences, labeled by acceptability (grammaticality) and divided into public (open) and test (hidden) parts. The public part of dataset contains 9,594 sentences and is divided, in turn, into training and development ("CoLA$_{dev}$") sets. The test set ("CoLA$_{test}$") contains 1,063 sentences with labels, hidden from the developer.

\textbf{Large Movie Review Dataset v1.0} ("IMDB") (\cite{maas-EtAl:2011:ACL-HLT2011}) contains 50,000 movie reviews, labeled by sentiment: "positive" or "negative". Labeled reviews are divided into two equal subsets, purposed for training and for testing. We applied additional lengths restriction to the samples of this dataset to obtain attention graphs of a reasonable size. Namely, we kept all reviews of size less than 128 tokens after tokenization with standard BERT uncased tokenizer ("Imdb$^{\le128}$"), and pruned away all others. After it, 5505 reviews remained in total. Then we divided the subset, suggested for testing purposes, into equal development and test sets.

\textbf{The SMS Spam Collection v.1} ("SPAM") (\cite{conf/doceng/AlmeidaHY11}) is a public set of SMS (text) labeled messages that have been collected for mobile phone spam research. It contains 5,574 real and non-encoded messages, tagged as legitimate (ham) or spam. For our purposes, we divided it into train, development and test ("SPAM$_{test}$") sets randomly in proportion $80 : 10 : 10$.

We used "development" subsets for tuning logistic regression hyperparameters: maximum amount of iterations and $l_2$-regularization coefficient. "Test" subsets were used for final validation. 

\subsection{Results}

\begin{table}[h!]
\centering
\begin{tabular}{||l|| c c c c ||} 
 \hline
  \rule{0pt}{2ex} & CoLA$_{dev}$ &  CoLA$_{test}$ & Imdb$^{\le 128}_{test}$ & SPAM$_{test}$ \\ [0ex] 
 \hline\hline 
 BERT \rule{0pt}{2.5ex} & 0.559 (82.0\%) & 0.492 & 0.833 (91.7\%) & 0.941 (98.7 \%) \\
 \hline
 $\beta_0, \beta_1, 144$ heads  & 0.549 (81.1 \%) & \textbf{0.508} & 0.812 (90.6 \%) & \textbf{0.950} (\textbf{98.9} \%) \\ 
 $\beta_0, \beta_1, 12$ best heads & 0.532 (80.7 \%) & 0.463 & 0.805 (90.3 \%) & 0.878 (97.3 \%)\\ 
 $\beta_0, \beta_1, 3$ best heads & 0.452 (77.1 \%) & 0.456 & 0.799 (90.0 \%) & 0.809 (96.1 \%)\\ 
 $\beta_0, \beta_1, 1$ best head & 0.427 (76.4 \%) & 0.385 & 0.735 (86.9\%) & 0.606 (92.3 \%) \\ 

\hline
 Test examples amount & 1043 & 1033 & 1415 & 556 \\
\hline
\end{tabular}
\smallskip
\caption{The comparison of our classification methods with the conventional BERT-based classifier by Matthew score and accuracy (in brackets, \%).}
\label{table_results}
\end{table}

As an efficiency measure of a linear classifier, we use Matthew score (Matthew coefficient), which is calculated by formula $$
\text{MCC} = \frac{ \mathit{TP} \times \mathit{TN} - \mathit{FP} \times \mathit{FN} } {\sqrt{ (\mathit{TP} + \mathit{FP}) ( \mathit{TP} + \mathit{FN} ) ( \mathit{TN} + \mathit{FP} ) ( \mathit{TN} + \mathit{FN} ) } },
$$
where we denote by $FP$, $FN$, $TN$, and $TP$
the amount of false positive, false negative, true negative, and true positive predictions of our classifier, respectively.
We also note classifications accuracy in brackets for those datasets, where test labels are available in open access.

For these experiments we fine-tuned BERT on each of datasets separately and used the set of six weight thresholds for calculating Betti numbers. In Table \ref{table_results} we emphasized in bold the results which surpassed the result of the fine-tuned BERT classifier.

For the first experiment we used the features calculated on all 144 heads. For consequent experiments, we checked Matthew score of classification upon features, built from the graph on each head on the train set, and ranged heads in descending order according to it. Then we picked 12, 3 or 1 heads with the best Matthew score and used them for calculation of classification features (Betti numbers) on development and test sets.

It's noticeable that topological features of attention graphs on particular heads have different linear separability. For more information about this 
see Appendix B.

\section{Conclusion and further research}
                                 
We have shown that the topology of attention graphs contains enough information for classifying texts by three different attributes: linguistic acceptability, sentiment, and being SPAM or not. Thus, we see here some degree of universality for distinguishing different text properties.

Moreover, the result of our linear classifier, trained on topological features, surpassed the result of the conventional BERT-based classifier on the hidden test subset of the Corpus of Linguistic Acceptability dataset and is a little better on the SMS Spam Collection v. 1 dataset. This allows us to suppose that these features may contain even more generalized task-relevant information
than the BERT output embedding. 
Plans of our future research include checking this daring statement 
with other topological features and other   threshold collections. 
Particularly, in our current work we didn't use the information about directions of graph edges, which could be utilized with directed graph invariants, such as number of simple directed cycles and number of strongly connected components of a digraph.

Another possible direction for future work is to use the information about differences between linear separability scores on different heads to determine which heads are more or less important for each particular task. Which can potentially be used as a base for new strategies of efficiently decreasing the model size.

\bibliographystyle{main}
\bibliography{references}

\section*{Appendix A. Persistent homology and Betti numbers}\label{appendix_topology}

Recall that a simplicial complex $K$ is a finite collection of finite sets called {\em simplices} such that each subset of any element of $K$ also is an element of $K$; such subsets of a simplex are called {\em faces}. 
In particular, an undirected graph is a simplicial complex where all edges and vertices are its faces.  
The set of all formal $\mathbb{Z}$-linear combinations of the $p$-dimensional  simplices (that is, $(p-1)$-element simplices) of $K$  is denoted $\mathsf{C}_{p}(K)$.  These linear combinations $c =\sum_{j} \gamma_{j}\sigma_{j}$ are called {\em $p$-chains}, where the $\gamma_j \in \mathbb{Z}$ and the $\sigma_{j}$ are $p$-simplices in $K$.

The boundary, $\partial(\sigma_j)$, is the formal sum of the $(p-1)$-dimensional faces of $\sigma_j$ and the boundary of the chain is obtained by extending $\partial$ linearly,
\begin{equation}\nonumber
\partial(c) = \sum_{j} \gamma_j\partial(\sigma_j),
\end{equation}
with integer coefficients $\gamma_j$.

The $p$-chains that have boundary $0$ are called $p$-{\em cycles}, they form a subgroup
$\mathsf{Z}_{p}(K)$ of $\mathsf{C}_{p}(K)$. 
The $p$-chains that are the boundary of $(p + 1)$-chains are called $p$-boundaries and form a subgroup $\mathsf{B}_{p}(K)$ of $\mathsf{C}_{p}(K)$.
The quotient group $\mathsf{H}_p(K) = \mathsf{C}_{p}(K) / \mathsf{B}_{p}(K)$ is called the $p$-th {\em homology} of $K$. 
Their ranks $\beta_p = \rk \mathsf{H}_p(K)$ of these abelian groups are called {\em Betti numbers}. The homologies and the Betti numbers 
are classical topological invariants of $K$.

In particular, a graph $G = (E,V)$ contains only 0-dimensional and 1-dimensional faces. It follows that its topological form is essentially described 
by the numbers $\beta_0$ and $\beta_1$ which are the only nonzero Betti numbers. Here $\beta_0$ is the number of connected components of $G$, and $\beta_1$ is the number of independent cycles of the graph (which is equal to $|E|-|V|+\beta_0$).

A {\em subcomplex} of $K$ is a subset of simplices that is closed under the face relation. A {\em filtration} of $K$ is a nested sequence of subcomplexes that starts with the empty complex and ends with the complete complex,
\begin{equation}\nonumber
\emptyset  \subset K_1 \subset K_2 \subset K_3  \subset \cdots \subset K_m = K.
\end{equation}
In particular, to any weighted undirected graph $G = (V,E)$ and an increasing sequence $0 = t_0 \le t_1 \le \dots \le t_m$ such that $t_m$ is greater or equal to the maximal edge weight in $G$, one can associate a filtration 
\begin{equation}
\label{eq:graph_filtration}
\emptyset   \subset G_{t_0} \subset \dots \subset G_{t_m} = G,
\end{equation}
where   $G_{t_i} = (V,E_{t_i})$ and $E_{t_i}$ consists of all edges of $E$ with weight more or equal to $t_i$.

The $p$-th persistent
homology of $K$ is the pair of sets of vector spaces $\{
\mathsf{H}_p(K_i) | 0\le i\le l \}$
and maps
$\{ 
f_{i,j}: \mathsf{H}_p(K_i) \to 
\mathsf{H}_p(K_i)| 1\le i < j \le l
\}$, where the maps are 
induced by the inclusion maps $K_i\to K_j$.

Each persistent homology class $\alpha$ in this sequence is ``born'' at some $K_i$ and ``dies'' at some $K_j$. One can visualize this as an interval $[i,j]$. The collection of all such intervals is called the {\em barcode} of the filtration. It is the most useful invariant of the filtration. Note that  the information about the persistent homology classes is generally essential to calculate the barcode, whereas the information about the Betti numbers only is insufficient.

Still, in the case of the filtration associated to a weighted graph~(\ref{eq:graph_filtration}),
the basis of $H_0$ (respectively, $H_1$) gives the intervals of the form $[0, t_i]$ (resp., $[t_i, t_m]$) only. 
Given a number $l=t_k$, the number of intervals of length at most $l$ for $H_0$
(respectively, the number of intervals of of length at least $t_m-l$   for $H_1$) is therefore equal to the 
the Betti number $\beta_0 (K_l)$ (resp., $\beta_1 (K_l)$). We see that in this case the collection of the  Betti numbers $\beta_i (K_{t_j})$ is sufficient to recover the barcode. Thus, we  
use just Betti numbers of the subgraphs $K_{t_j}$ as the only topological invariants of our graphs.

\newpage

\section*{Appendix B. Classifiers built by a single head}\label{appendix_heads}

\begin{figure}[h]
\centering
\includegraphics[width=0.99\linewidth]{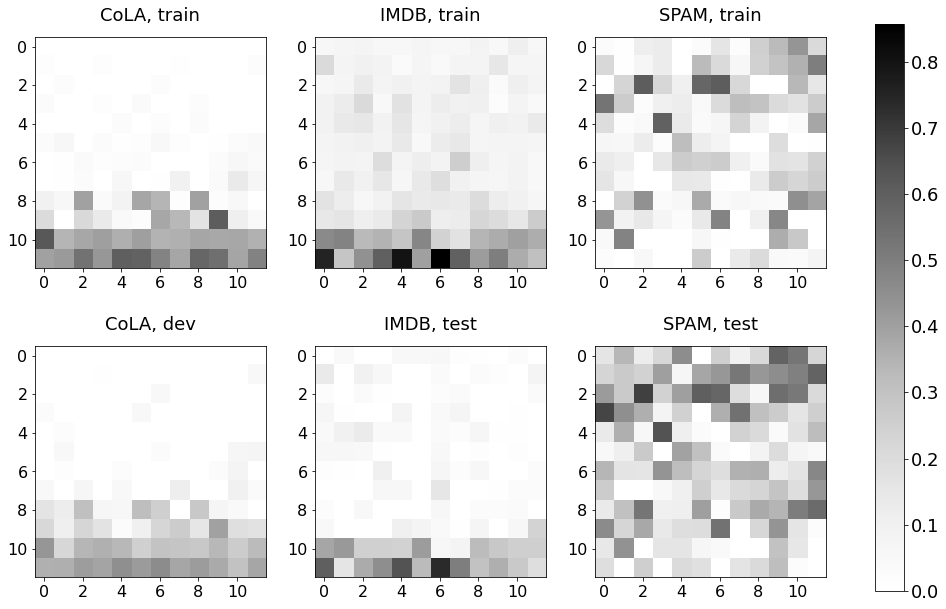}
\smallskip
\caption{Matthew scores of predictions of linear classifiers, 
built upon particular attention heads. 
The number of the layer is displayed on the vertical axis. 
The number of the head inside the layer is displayed on the horizontal axis. \\}
\label{heads_small}
\end{figure}

Figure \ref{heads_small} illustrates that the relevance of features, calculated on different heads, varies greatly from head to head on each task. It also shows that the same head can be more relevant for solving one task but less relevant for solving other ones. On the other hand, we can see similar patterns on the train and test/development sets for each task separately (in each column of Figure \ref{heads_small}). This means that the head importance, derived from this score, is generalized to unseen examples and therefore can be used for feature selection.

\end{document}